\pdfoutput=1

\documentclass[11pt]{article}

\usepackage{EACL2023}
\usepackage{times}
\usepackage{latexsym}

\usepackage[T1]{fontenc}

\usepackage[utf8]{inputenc}

\usepackage{microtype}
\usepackage{multirow}
\usepackage{booktabs}
\usepackage{amssymb}
\usepackage{amsmath}
\usepackage{graphicx}
\usepackage{mathrsfs}
\usepackage{float}

\newcommand{\CDS}{\protect{ControversialQA} } 

\title{ControversialQA: Exploring Controversy in Question Answering}

\author{Zhen Wang$^{*1}$, Peide Zhu$^{*2}$, Jie Yang$^{3}$ \\
  $^1$Tokyo Institute of Technology, $^{2,3}$Delft University of Technology \\
  $^1$wzh@lr.pi.titech.ac.jp, \{$^2$p.zhu-1, $^3$j.yang-3\}@tudelft.nl \\
  }

\begin{document}
\maketitle
\begin{abstract}

Controversy is widespread online. Previous studies mainly define controversy based on vague assumptions of its relation to sentiment such as hate speech and offensive words. 
This paper introduces the first question-answering dataset that defines content controversy by  user perception, i.e., votes from plenty of users. 
It contains nearly 10K questions, and each question has a best answer and a most controversial answer.
Experimental results reveal that controversy detection in question answering is essential and challenging, and there is no strong correlation between controversy and sentiment tasks.
We also show that controversial answer and most acceptable answers cannot be distinguished by retrieval-based QA models, which may cause controversy issues.
With these insights, we believe \CDS can inspire future research in controversy in QA systems.




\def\thefootnote{*}\footnotetext{Equal Contribution}\def\thefootnote{\arabic{footnote}}

\end{abstract}

\section{Introduction}
\label{sec:intro}
Large numbers of people are participating in online discussions on a daily basis, asking and answering questions, or expressing their opinions on certain topics, on various platforms such as Twitter, Facebook, and Reddit. 
It is common that the discussions may turn in opposite directions, and controversy may arise when people have conflicting opinions, especially in political, health or entertainment topics.
Online platforms need to detect controversy in these discussions since the controversial texts may contain fake information or some user groups may find them inappropriate, offensive, or unwanted. 

Early research on controversy in social media mainly focuses on political activities, especially the president election, for the purpose of predicting the election result \cite{politicalblog, partisanpolitical}. 
\citet{roleoftwitter} extends the controversy research to a broader area, using a set of controversial topics on Twitter to investigate user behavior. They find that Twitter tends to relay one person's views to others who hold similar views. \citet{newscontroversy} find that there is more negative effect and biased languages in readers discussions on controversial online news.  
While demonstrating the existence and the potential negative impact of controversy, these previous works define controversy using syntactic heuristics, i.e., 
characterising controversy as a consequence of hate speech or offensive words.
This is not necessarily true since the controversy is usually seen as a property of the semantic, e.g., the conflicting opinions among people, and sometimes can only be judged by user perception.
Therefore, in this paper, we introduce \textbf{\emph{ControversialQA}}, the \emph{first-of-its-kind} dataset collected focusing on the controversial discussions, where the controversy is defined by user perception (votes from plenty of users). We specifically focus on content in question answering (QA) since QA forums in Reddit tend to inspire discussions from people with various opinions, therefore, controversy results in both the question and the answer.

Based on \CDS, we first propose a series controversy detection tasks and show it is a novel and challenging dataset. 
Second, in contrary to assumptions in previous works, we reveal that controversy detection is not firmly correlated to existing sentiment analysis or hate speech detection tasks.
Further, our experiments on \CDS prove the current retrieval-based QA model may be opposed by controversy issues, i.e. instead of providing the most acceptable answer, it may choose answers that can lead to opinion split.


\section{Dataset}
\label{sec:ds}
\subsection{Data Collection}

\begin{figure}[t]
  \centering
  \includegraphics[width=0.8\hsize]{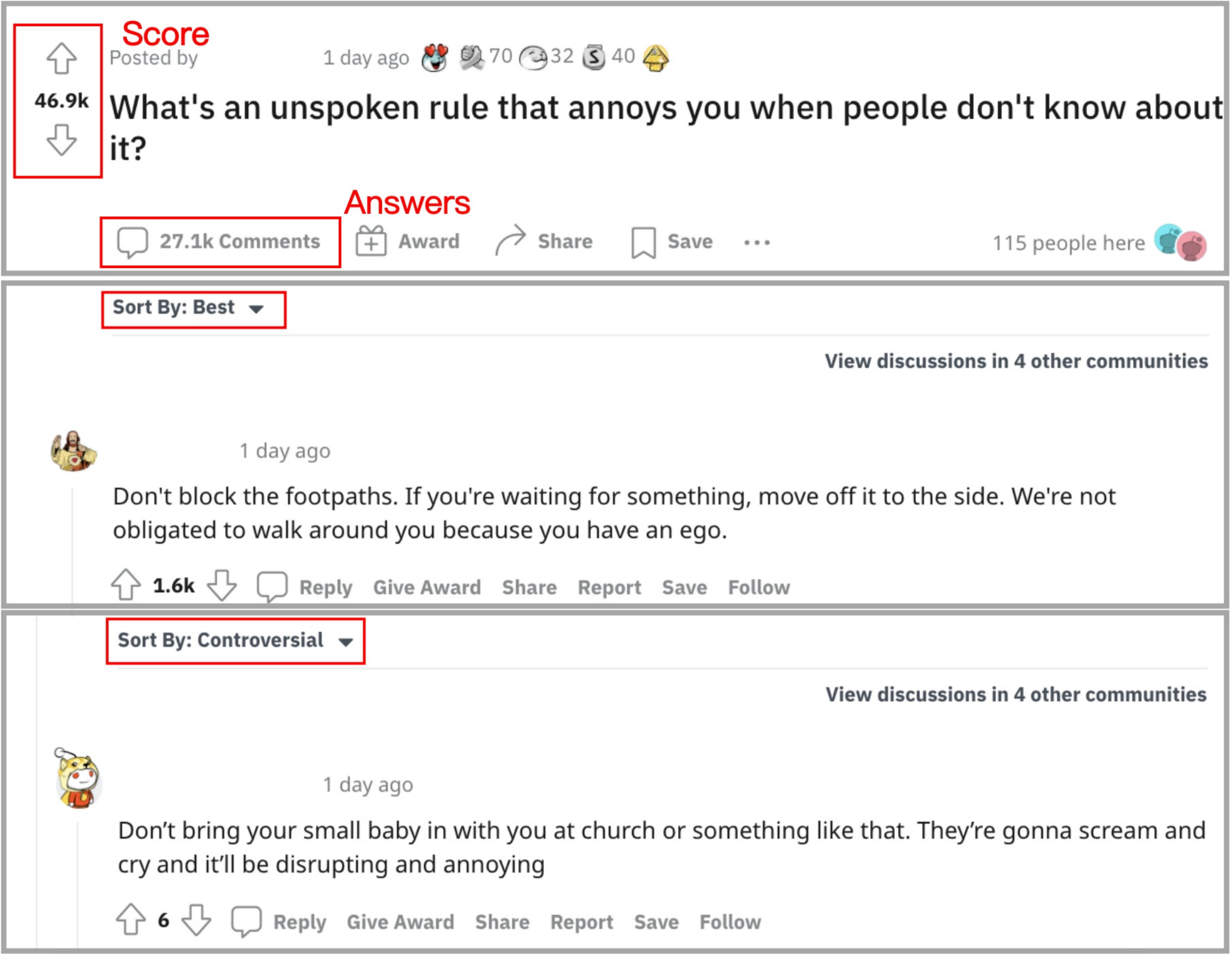}
  \caption{An sample from r/AskReddit. The upmost two red boxes represent the question score and the current number of answers. The other two show a top ranked answer using \textit{sort by best} and \textit{sort by controversial}.}
  \label{fig:qasample}
\end{figure}

Since we focus on controversy detection in online open QA, we chose the \textit{r/AskReddit} thread in Reddit~\footnote{\url{https://www.reddit.com/r/AskReddit/}} as the data source. 
\textit{r/AskReddit} is a subreddit where users can ask questions or answer questions in other posts.
With the help of API~\footnote{\url{https://www.reddit.com/prefs/apps}} provided by Reddit, we collect questions that have been posted to \textit{r/AskReddit} over the past ten years from 2012 to 2021. 
For each question, we collect the question itself and a most controversial answer; for comparison, we also collect a best answer which is supported by most of readers.
A sample can be found in Figure \ref{fig:qasample}. The first two red boxes represent the question score (the difference between upvotes and downvotes that users vote for this question) corresponding to the question quality and the current number of answers for the question. 

The default sorting formula in Reddit is \textit{sort by best}, where the larger the difference between the number of \textit{upvotes} and \textit{downvotes}, the higher the answer ranks. Another sorting formula is \textit{sort by controversial}, where an answer with a higher controversial score will be listed ahead. The formula~\footnote{\url{https://github.com/reddit-archive/reddit/blob/master/r2/r2/lib/db/_sorts.pyx}} Reddit uses to calculate the controversial score is shown in Equation~\ref{eq:score} and the function graph is in depicted Figure \ref{fig:func}.

\begin{align}
\begin{split}
    & \text { magnitude } = ups + downs \\ 
    & \text{ balance } = 
        \begin{cases} 
            \frac{downs}{ups}, & ups > downs; \\
            \frac{ups}{downs}, & ups \leq downs.
        \end{cases}  \\ 
    & \text{ controversial\_score } = magnitude^{balance} 
\end{split}
\label{eq:score}
\end{align}
where \(ups\) is the number of ``upvote'' and \(downs\) is the number of ``downvote''. This algorithm ranks an answer to be more controversial when having both large numbers of upvotes and downvotes.

\subsection{Quality Control}

To ensure the quality of the dataset, i.e., to ensure that a controversial answer indeed receives upvotes and downvotes by a large number of users, we only retain questions with more than 100 answers and more than 100 question score. Since answers on Reddit sometimes come up with meaningless words or phrases such as ``no'' and ``knock knock'', only samples whose answers contain more than 20 words are retained. 
To make the distribution of the answer length more balanced, we also delete samples with too long answers if they contain more than 150 words. The statistics of ControversialQA are shown in Table \ref{tab:statistics}.

\section{Experiments And Analysis}
\label{sec:exp}

\begin{table}[]
\centering
\small
\begin{tabular}{ll}
\toprule
\textbf{Metric}                                   & \textbf{Counts} \\\hline
\#Total Instance                         & 9952   \\
\#Average tokens of question             & 15.58  \\
\#Average tokens of best answer          & 58.49 \\
\#Average tokens of controversial answer & 52.99 \\
\bottomrule
\end{tabular}
\caption{Statistics of ControversialQA.}
\label{tab:statistics}
\end{table}

\subsection{Experimental Settings}

We randomly split the dataset into training/validation/testing sets in the ratio of 8:1:1 for all the tasks.
We conduct experiments on different tasks using BERT \cite{bert} and RoBERTa \cite{roberta} models provided by Hugginface\footnote{\url{https://github.com/huggingface/transformers}}, since these pre-trained language models have achieved state-of-the-art results on various text classification tasks.
To make the QA pairs conform to the input format of BERT, we use \texttt{[SEP]} to concatenate the answer and question to `` \texttt{[CLS]} Answer \texttt{[SEP]} Question''.
The output of the \texttt{[CLS]} token is used for the linear classification tasks.
We use F1 score as the evaluation metric. 
For each training process, we train the model with the training set, retain the model that performs best on the validation set, and apply it on the testing set. 
Adam \cite{adam} is used as the optimizer and the learning rate is 1e-5.
Cross Entropy is used as the loss function.

\subsection{Controversy Detection on ControversialQA}

We conduct two tasks on ControversialQA: 
\emph{\textbf{(I) Controversy Classification Task:}} a QA pair is used as the input, and the output is a bool value representing whether this answer will cause controversy for the current question. 
\emph{\textbf{(II) Controversy Selection Task:}} given two candidates, a model should correctly select which answer will cause controversy. 
The training process is the same as that of controversy classification. 
During the test, for each sample, we use softmax to calculate the score of the two answers respectively and take the one with the higher score as the controversial answer.

The experimental results are shown in Table \ref{tab:experiments}.
Combining the two experiments, we get three major findings. 
(1) RoBERTa-large achieves best performance in both tasks, which achieves the highest at 74.83 and 86.84 in F1 respectively, 
showing that current methods can achieve relatively good results, yet there is still a lot of room for improvement.
(2) Directly detecting controversy of an answer is more difficult than comparing the controversy of two answers (\textbf{74.83} vs. \textbf{86.84}); 
(3) For both tasks, the large version of the model performs better than the base version, and the better pre-trained model performs better (RoBERTa better than BERT), proving that parameters and pre-training are helpful to improve the accuracy in detecting controversy.

\begin{table}[]
\centering
\small
\begin{tabular}{lccc}
\toprule
\multicolumn{1}{l}{\textbf{Model}} & \textbf{AR}  & \textbf{AM} & \textbf{AW} \\ \hline
\multicolumn{4}{c}{\textbf{Controversy Classification}}             \\ \hline
BERT-base                 & 68.65 & 65.29  & 62.88    \\
BERT-large                & 70.01 & 60.99  & 63.24    \\
RoBERTa-base              & 71.91 & 67.21  & 65.17    \\
\textit{+sentiment}                & 68.51 & 66.20  & 62.44    \\
\textit{+offensive}                & 69.83 & 66.76  & 64.26    \\
\textit{+irony}                    & 70.18 & 65.75  & 63.05    \\
\textit{+hate}                     & 70.17 & 65.32  & 65.06    \\
\textit{+emotion}                  & 70.51 & 66.50  & 62.78    \\
RoBERTa-large             & \textbf{74.83} & \textbf{69.96}  & \textbf{66.50}    \\ \hline

\multicolumn{4}{c}{\textbf{Controversy Selection}}                      \\ \hline
BERT-base                 & 78.51 & 72.12  & 70.19    \\
BERT-large                & 80.12 & 74.01  & 72.02    \\
RoBERTa-base              & 83.53 & 76.15  & 73.80    \\
\textit{+sentiment}               & 82.12 & 74.65  & 72.38    \\
\textit{+offensive}                & 80.92 & 74.26  & 71.69    \\
\textit{+irony}                    & 80.82 & 73.94  & 71.15    \\
\textit{+hate}                     & 81.32 & 74.57  & 73.14    \\
\textit{+emotion}                  & 81.93 & 75.36  & 72.76    \\
RoBERTa-large             & \textbf{86.84} & \textbf{77.80}  & \textbf{75.99}    \\ 
\bottomrule
\end{tabular}
\caption{F1 score of controversy classification and choice. \textbf{AR} means \textit{r/AskReddit}, \textbf{AM} means \textit{r/AskMen}, \textbf{AW} mean \textit{r/AskWomen} (Section \ref{sec:ood}).}
\label{tab:experiments}
\end{table}

\begin{table*}[]
\centering
\small
\begin{tabular}{p{0.14\textwidth}p{0.40\textwidth}p{0.38\textwidth}}
\toprule
\textbf{Question} & \textbf{Choice A} & \textbf{Choice B} \\\hline
    What is one thing people do NOT want for christmas?     &   \textcolor[rgb]{0.09, 0.45, 0.27}{Gift cards. Don't force me to spend virtual money at a store I don't shop at, only to have to fork over a few bucks at the register to keep there from being a few bucks left on the card. We prefer cash...}
       &    When we were younger my brother and I thought it would be funny as hell to give our little sister coal for Christmas ... of course she cried her eyes out so I'm gonna go with coal if you're a kid.
     \\\hline
    What TV series isn't worth finishing?     &    That 70s show. Generally people say that Randy was the reason for the shitty ending, but IMO the show shouldn't have tried to change the original ...      &   \textcolor[rgb]{0.09, 0.45, 0.27}{I could never get through Breaking Bad. ive tried a few times and just cant. I cant get last season 4 it just gets so slow and boring} \\\hline
    What is one conspiracy that you firmly believe in? and why?	& \textcolor[rgb]{0.09, 0.45, 0.27}{9/11 was an inside job. If you examine the evidence for an inside job theory vs. the evidence available for the Bush administration's theory it is overwhelmingly in the favor of an inside job...} & Cinnabon vents their oven exhaust directly into the food court to increase sales. Can you smell the food from the Sbarro, Panda Express or McDonalds in the food court? Of course not... \\
\bottomrule
\end{tabular}
\caption{Three error predicted samples by fine-tuned RoBERTa-large. The choice with \textcolor[rgb]{0.09, 0.45, 0.27}{green} color is the ground-truth answer that will cause controversy.}
\label{tab:erroranalysis}
\end{table*}

\subsubsection{Case Study}

In Table \ref{tab:erroranalysis}, we present three samples that are wrongly classified in the controversy selection task. 
In the first question, the description of gift cards leads to controversy. Some people think gift cards are helpful because the cards allow more freedom to buy gifts, while some don't like gift cards because the cards force them to buy things in a store they don't like. 
In the second question, the judgment to \textit{Breaking Bad Season 4} causes controversy. Some people think the season is interesting, while some others hold the the opposite attitude as the one providing the given answer. 
As for the last question, many of the 9/11 conspiracy theories have already been disproved, but some people still believe in them. 

Those samples reveal that controversy in the real world is caused by various factors such as personal experience, political opinions, etc., that can often go beyond the text itself.
Therefore, to achieve better controversy detection, the model should not only consider the answer text, but also  consider contextual, commonsense, or external knowledge. 

\subsubsection{Out of Domain Test}
\label{sec:ood}

Since \textit{r/AskReddit} is an open domain topic, we want to figure out whether the model trained on it can be migrated to other specific domain topics. Therefore, we collect two auxiliary datasets with similar size of the ControversialQA testing set from \textit{r/AskMen} and \textit{r/AskWomen}  using the same collecting procedure and then apply the trained models to them. 
\textit{r/AskMen} primarily aims at male Reddit users, asking questions about men and expecting men to answer them, while \textit{r/AskWomen} is just the opposite. 
Experiment results are shown in the \textbf{AM} and \textbf{AW} columns of the Table \ref{tab:experiments}.

A huge gap between in and out of domain performance can be observed.
Apparently, gender has a significant impact on discussion subjects and opinions.
This shows that although \textit{r/AskReddit} is an open domain topic, containing a variety of questions and involving all genders, the same result cannot be achieved if the model is applied directly to a specific domain, which implies that controversy detection performance is closely related to user demographics.



\subsection{Can Sentiment Help Detect Controversy?}
Sentiment analysis is usually involved in previous controversy studies.
To investigate whether the controversy detection task overlaps with sentiment analysis, 
first, instead of original RoBERTa, we perform controversy detection by fine-tuning TweetEval~\cite{tweeteval}, a collection of RoBERTa-base models pre-trained on several different kinds of sentiment datasets, including \textit{sentiment}, \textit{offensive}, \textit{irony}, \textit{hate}, \textit{emotion}, on ControversialQA. 
The results are shown in Table \ref{tab:experiments}. 
We then label QA pairs by both those sentiment models and humans. Human sentiment annotation is conducted with three experts and harvests 500 labeled pairs by simple major voting. 
We finally calculate their correlation with ground-truth controversy labels.
The results are shown in Figure \ref{fig:correlation}. 
Experimental results show that i) pre-training on sentiment datasets cannot help to improve the performance on controversy tasks; ii) neither automatic nor human sentiment labeling show strong correlation between previous sentiment tasks and ControversialQA. 

\subsection{How Does Controversy Influence Retrieval-based QA Models?}
Compared to controversial answers which may contain fake information or views opposed by some user groups, the best answer should always be the first choice of IR-based QA model.
To figure out whether current QA models can evaluate the quality of the answers from the controversial perspective, we use a DistilBert~\cite{hofstatter2021efficiently} retriever pre-trained on MS MARCO~\cite{msmarco} to calculate the score between the question and two answers separately, where one is the best answer which is supported by majority of users and the other is the controversial answer. 
We do statistics across the entire dataset, and the result shows that there is a \textbf{52.43\%} (ideal is 0\%) chance that model regards the controversial answer more relevant (better) than the best one. 
This proves that current QA models which rank answers merely rely on the relevance between question and answers may raise controversy issues. 
This can cause an answer that appears fitting the question but actually contain some controversial information, such as fake news or gender discrimination, ranking at the top, which is unacceptable for real-world scenarios.
Our dataset can be used to study how to improve the appropriateness and credibility of answers.

\section{Related Work}
\label{sec:rw}

With the development of social media, controversy now spreads over everywhere on the Web. 
However, there are only limited resources for the research of controversy in online discussion platforms; most of them are based on Twitter~\cite{whatsyourevidence, controversialdebates}.
Some researches regard debates over twitter topics as the source of controversy~\cite{whatsyourevidence, controversialdebates}. \citet{ebbandflow} study the ebb and flow of controversial twitter debates in certain topics. The research from \citet{impact} shows that disinformation are pervasively existing in debates. Lots of previous controversy-related researches assume controversy is strongly connected with sentiment~\cite{roleoftwitter, newscontroversy}.
Sentiment analysis~\cite{sentimentanalysis} is the study of identifying the sentiment expressed from a given text. 
However, these previous definitions is not in line with the connotation of controversy.

\section{Conclusion}
\label{sec:conclusion}
In this paper, we present \CDS,  the first dataset used for controversy detection in QA. Then we conduct comprehensive experiments on this dataset, which show that controversy detection is a challenging task and has no strong correlation to sentiment tasks. We further prove that retrieval-based QA models cannot distinguish between the most acceptable answer and controversial answer semantically and may be opposed by controversy issues in real-world scenario.
These insights we believe can inspire future research in several fields.


\bibliography{anthology,custom}
\bibliographystyle{acl_natbib}

\appendix
\section{Appendix}
\label{sec:appendix}
\subsection{Controversial Score Function}

\begin{figure}[H]
  \centering
  \includegraphics[width=0.8\hsize]{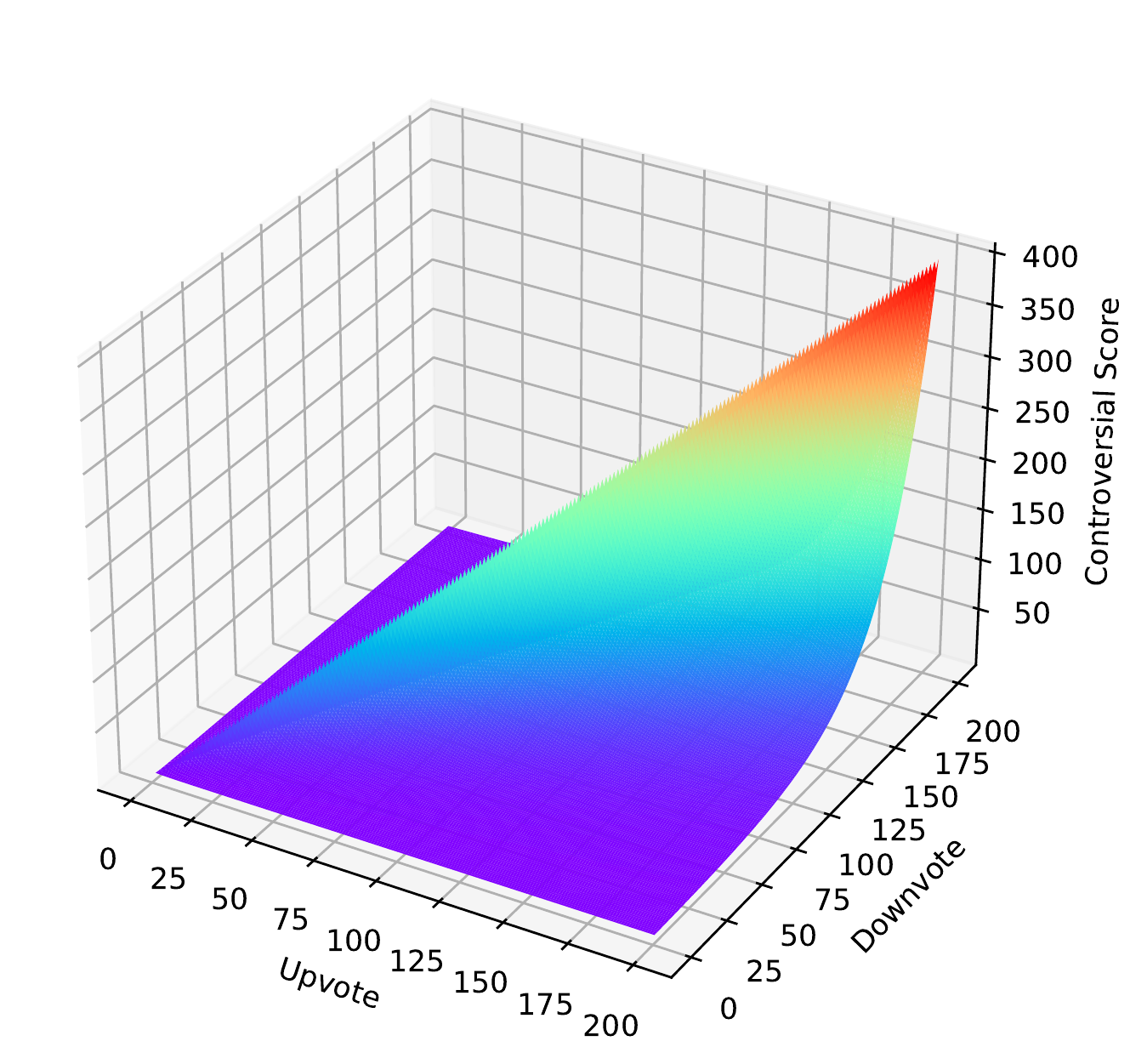}
  \caption{Function graph of controversial score algorithm.}
  \label{fig:func}
\end{figure}

In this section we draw the function graph of controversial score algorithm in Figure \ref{fig:func}. It is an axially symmetric function along the diagonal. The value of the controversial score is maximum when the values of the upvote and downvote are equal, and controversial score increases as the values of upvote and downvote increase.

\subsection{Correlation Between Controversy and Sentiments}
\label{sec:correlation}

\begin{figure}[H]
  \centering
  \includegraphics[width=1.0\hsize]{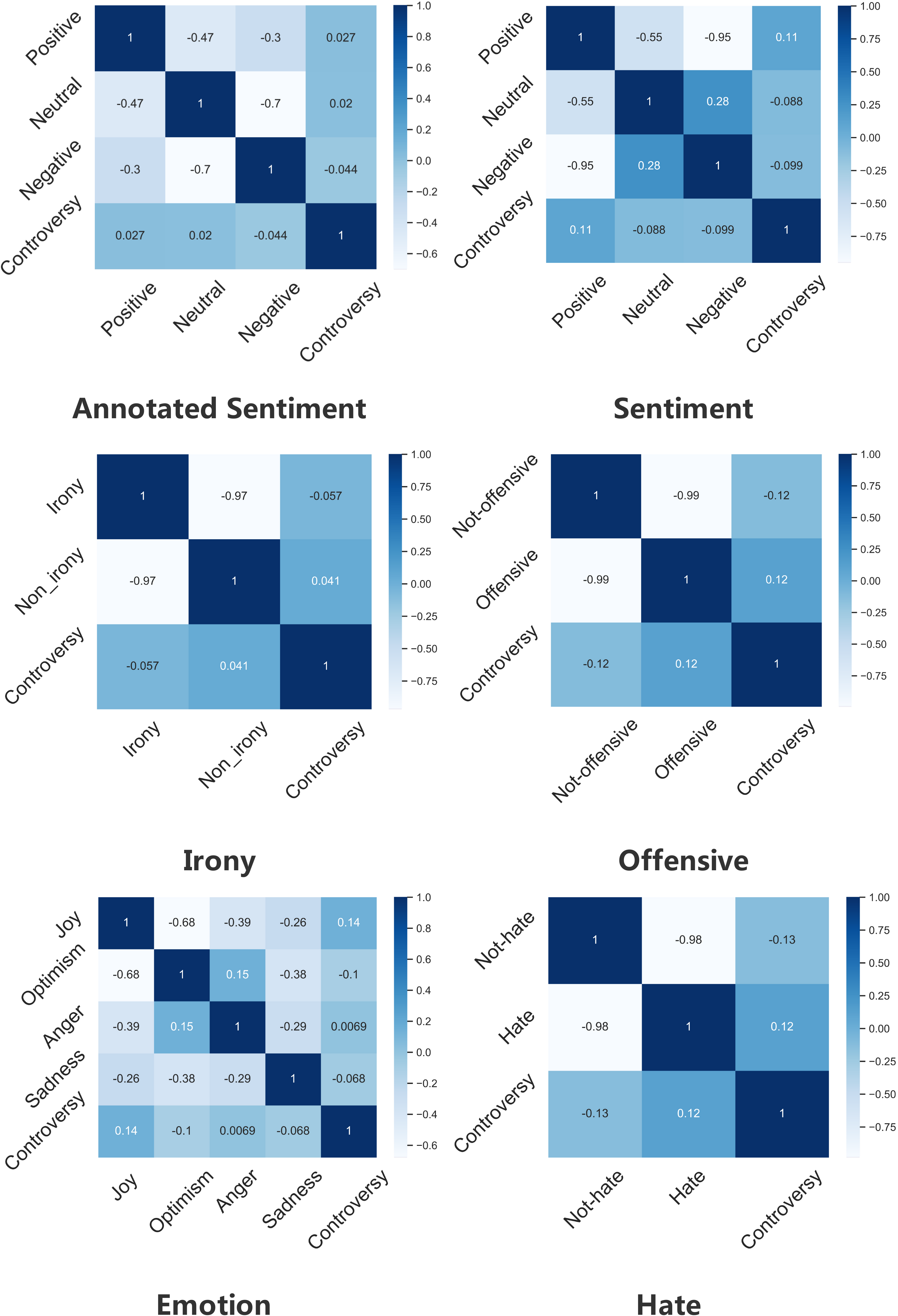}
  \caption{Correlation between \textbf{controversy} and different types of \textbf{sentiments}.}
  \label{fig:correlation}
\end{figure}

In this section we provide the correlation maps between controversy and different kinds of sentiment in Figure \ref{fig:correlation}. The first ``Annotated Sentiment'' map is calculated by the correlation between human annotated sentiment labels and ground-truth controversy labels from ControversialQA. And the other five maps is calculated by the correlation between the labels generated by pre-trained TweetEval models and ground-truth controversy labels.
It can be observed that there is no strong correlation between those sentiments and controversy.

\end{document}